\documentclass[11pt]{article}
\usepackage{eacl2017}
\usepackage{times}
\usepackage{url}
\usepackage{latexsym}

\eaclfinalcopy 
\def\eaclpaperid{472} 

\title{Age Group Classification with Speech and Metadata Multimodality Fusion}

\author{Denys Katerenchuk\footnotemark  \\ 
  CUNY Graduate Center \\
  365 Fifth Avenue, Room 4319\\
  New York, USA\\
  {\tt dkaterenchuk@gradcenter.cuny.edu}}

\date{}

\begin{document}
\maketitle
\begin{abstract}
  Children comprise a significant proportion of TV viewers and it is worthwhile to customize the experience for them. However, identifying who is a child in the audience can be a challenging task. 
We present initial studies of a novel method which combines utterances with user metadata. In particular, we develop an ensemble of different machine learning techniques on different subsets of data to improve child detection. Our initial results show a 9.2\% absolute improvement over the baseline, leading to a state-of-the-art performance. 

\end{abstract}

\footnotetext[1]{This work was done while the author was an intern at Comcast Research.}

\section{Introduction}
\label{sec:intro}

Building on recent breakthroughs on speech understanding, people ask their cellphones any questions and expect to get reasonable answers, or ask their TVs for movie recommendations. 
The identity of the user plays a key role in personalizing and improving these actions. For instance, in the case of movie request, a general, probabilistic model will not work well. 
 Consider a case when a child asks to watch "Ruby and Max," an animated television series, but the automatic speech recognition system (ASR) mistakenly resolves it to the popular "Mad Max" movie in the downstream natural language processing (NLP) module. Having the knowledge of the age, the system could fix such errors by returning age-relevant results. 

Unfortunately, this scenario is quite common considering that even state-of-the-art ASR systems produce very bad results on understanding children's speech. There are a couple reasons for this: 1) most ASR systems are trained to understand adults, 2) children's voices are hard to analyze because of not fully developed vocal tracts \cite{shivakumar2014improving}.
One way to improve the performance is to add an intermediate system that can identify users.


In this paper we investigate child identification from voice commands, metadata and the combination of the two to improve classification accuracy. Age and gender identification from speech is not a new problem and much research has been done in this area (sec.\ref{sec:relatedwork}), yet the results are far from perfect. In particular, the task to identify adults from kids becomes more challenging when the utterances are only a couple of seconds long. 
We investigate a novel multimodel approach to improve classifier accuracy by combining speech data with rich usage metadata (sec.\ref{sec:data}). Specifically, we extract features separately from speech and usage data, and build individual models that are fused together (sec.\ref{sec:methods}) to improve classifier performance. The results are described in section \ref{sec:results}. 


\section{Related Work}
\label{sec:relatedwork}
Speaker information, such as accent, gender or age, can be used to improve speech understanding \cite{abdulla2001improving}, provide background information, and advance human-computer interactions. A human vocal tract undergoes changes starting from birth and continues throughout one's life.  \newcite{brown1991speaking} found that fundamental frequencies directly correspond to the ages of professional singers. Later, \newcite{naini2006speaker} investigated the correspondence of MFCCs, shimmer, and jitter to a speaker's age. They found that jitter and shimmer do, indeed, help distinguish ages, but only on wider age rages. With application of more advanced machine learning techniques, \newcite{metze2007comparison} achieved human level performance on longer speech segments, while short utterances were challenging to classify correctly. The recent work on a similar task of gender identification by \newcite{levitan2016identifying}, revealed that human level performance is achievable on short utterances as well. 

In this work, we build on the prominent research approach, and investigate its performance on a challenging real world data set: TV domain where utterances are only about a second long.  This is why in addition to speech, we analyze metadata, which is commonly ignored, to explore a fusion of multiple models in the classification task. We compare the performance of three models based on SVM, random forest, and deep learning, then report the results.


 





\section{Data}
\label{sec:data}

The speech data is collected at random each week for over a year's time span and was manually labeled by human annotators as "MALE", "FEMALE", or "KID". Since we don't have ground truth labels, we use these labels as the gold standard. "MALE" and "FEMALE" labels are combined into one "ADULT" class. Each audio is a short, on average 1.2 seconds long, command from a user to a TV box such as "watch SpongeBob" or "CNN." In total we have 15,001 instances of labeled utterances where 3,848 were labeled as "KID". To normalize the data set, we at random sampled 3,848 utterances with the "ADULT" label. This is done to create a balanced dataset of 7,696 instances. 
The data was split into train and test sets with 75:25 ratio leaving 5,772 for training and 1,924 instances for testing sets. Cross validation on the train set was used as our development set to optimize the algorithms.
The final results are reported on the test set.

In addition to the voice commands, we collected user metadata. This data contains general usage patterns such as date, time, and expected audience type (children or adults) of the requested TV show. The data covers only one month of activity which makes the data meager. As a result, we ignore dates and use weekdays instead. Additionally, we calculate the likelihood of a request made for a children's show in a given weekday and hour. All the times and date were converted to the user local time zones. In addition, we use a hand written rule that treats all commands between 11pm and 6am as commands from adults. The reason is that most of the time children are in bed during these hours and in way we are eliminating false positives. 


%

\section{Methods}
\label{sec:methods}
\subsection{Feature Extraction}
Before the feature extraction step, the audio was preprocessed and normalized. In the preprocessing step, all silences were removed to keep user commands. In the normalization step, we try to mitigate variance in speech by normalizing the volume. This is a common preprocessing step that is used in ASR systems. After these two preprocessing steps, we extract features to use as an input to train an acoustic model. In order to validate the quality of the preprocessing steps, acoustic features are extracted from raw, preprocessed, and normalized audio.

For acoustic feature extraction we use the open-source tool openSMILE \cite{eyben2010opensmile}. OpenSMILE is a well known utility that produces state-of-the-art acoustic features and often used during annual INTERSPEECH paralinguistic challenges to define a baseline. The source code includes a set of configuration files for different features. The configuration file we use in our experiments is "paraling\_IS10.conf." This version was introduced during the INTERSPEECH 2010 Paralinguistic Challenge \cite{schuller2010interspeech}. The challenge was to create predictive models for gender and age classification. We also tried to experiment with other configuration files; however, showed lower performance.

We extract 1582 acoustic features from each user utterance. 
The features are created by first extracting low level descriptors (LLDs) of 10ms frame level step and 20ms window size. The LLDs include a total of 34 features such as 12 MFCCs, F0, energy, jitter, etc. After that, we derive 34 deltas from the LLDs and apply a set of 21 functions. A list of the functions is shown in table \ref{table:1} and complete feature description can be found in \cite{schuller2010interspeech}.

\begin{table}
	\centering
	\begin{tabular}{| c | c |}
	\hline
	LLDs & Functions \\
	\hline \hline
	mfcc 0-14 & mean/max/min Pos \\
	pcm\_loudness & linregc 1 2 \\
	logMelFreqBand 0-7 & linregrr A Q\\
	lspFreq from 8 LPC & stddev, kurtosis \\
	F0finEnv & quartile 1,2,3\\
	voicingFinalUnclipped & percentile 1, 99 \\
	F0 & prtl\_range 0,1 \\
	jitter L/DDP & \\ 
	shimmer & \\
	\hline 	
	\end{tabular}
	\caption{Acoustic features}
	\label{table:1}	
\end{table}

In addition to speech, we explore the user requests. Despite ever-changing TV content, some phrases or words can aid to identify the viewer's age. We use an ASR system on each utterance to extract a transcript. Since the commands are very short and specific to the domain, a simple bag-of-word language model \cite{zhang2010understanding} is sufficient. From the dictionary of 5092 unique words, 2000 of the most frequent words are used as a word feature vector.


For each utterance, we also use its metadata such as weekday and hour. The intuition for including this data is that some TV content is targeted for a particular audience with respect to the time of the day. For example, news tend to run during evening hours and children oriented shows are shown in the morning or during a day.

In addition, we use the show-type request distribution from a given device. The distribution is derived by computing the percentage of children's shows against all shows watched during the specific time. 
We derive this distribution as a score from 0 to 1 for each hour, day, and entire one month time period for a given device. 
If a command comes between 11pm and 6am, we mark it with 0, assuming that only adults can be awake during these hours. 
As a result, two feature datasets were created: 1) time usage data that contains usage frequency per hour and weekday as well as content type, 2) ratio usage data that includes distribution of kids and non-kids content
requested from a given device. The ratio is calculated for each hour, day, and a given device in general. Considering that usage patterns of users with and without children vary, these datasets will add important information to make better classification decisions.

\subsection{Classification}
For classification, we use two well known algorithms: support vector machines (SVM) \cite{suykens1999least} and random forest \cite{liaw2002classification}. Both algorithms show state-of-the-art performances in speaker classification tasks \cite{Ahmad2015GenderIU}.
SVM algorithm works by creating support vectors that separating two classes in n-dimensional space, where each dimension is represented by a feature. The separation is done by finding the largest separation margin between the features from the two classes and a vector. Random forest is a tree based ensemble algorithm that work by running multiple decision tree algorithms, which are known as weak learners, at the same time. Each algorithm at random selects features and makes its decisions. At the end, all the results from the each learner are combined to provide the prediction. 
The models are trained using scikit-learn toolkit \cite{scikit-learn}, an open-source machine learning library. Both algorithms were used for training. However, only the best algorithm is used on the test data. 

Deep learning \cite{lecun2015deep} has shown to be a useful technique in many areas including audio processing. We build a deep network classifier with four hidden layers. Each layer is fully connected with a 50\% dropout rate to reduce overfitting \cite{srivastava2014dropout} and a sigmoid activation function \cite{marreiros2008population}. The last layer uses softmax activation and 0\% dropout rate. The size of each layer is chosen to first generalize the features and then narrow the feature space size. The best architecture has the following layer sizes [1582, 1582*8, 2048, 512, 64, 2]. The first and last layers are acoustic feature input and predicted binary class output respectively. The network is trained overnight on a consumer level GPU. 

During the training, we start with audio preprocessing by applying energy normalization and silence removal techniques. While energy normalization is a useful method to improve ASR performance \cite{li2001robust}, the results need to be tested to determine if this approach is applicable to our task. The removal of silences, on the other hand, is a valid step to increase the accuracy. After determining the best audio normalization, we train a separate model for each feature set: 1) audio, 2) time usage data, and 3) show-type request distribution. The models are tested with cross validation and the scores are reported in section \ref{sec:results}.
The test sets are used only at the end to evaluate the models on previously unseen data. In this way we avoid overfitting by tuning the algorithms on train data with cross validation that we use as development dataset.
At this point we have three datasets, which are acoustic data, time usage, and content type ratio. 
Each model is evaluated separately on the corresponding dataset.

Leveraging multi-domain data, we apply feature and model level fusion methods.
We experiment with combining features from the three domains into a single feature vector and train additional model on these features. At the same time, we perform model level fusion \cite{huang2011speaker}. Each trained model's output probability is used as inputs to AdaBoost ensemble learning algorithm \cite{ratsch2001soft}. We apply this approach only on the test data. The evaluation is done using cross validation on the test dataset.

\section{Results}
\label{sec:results}

\subsection{Baseline}
For the baseline, we use INTERSPEECH 2010 paralinguistic gender and age challenge's  pipeline \cite{schuller2010interspeech}. 
The data that was used for the challenge is different from ours in terms of audio domain, quality, and utterances were on average 2.2\% longer. Longer audio segment provide more information making the task easier. Further more, the challenge was to classify users of 4 age groups while we perform binary classification.  For this reason we cannot directly compare the scores. 
However, we follow the steps to replicate the challenge's pipeline on our unaltered data and use the score as our baseline. The accuracy of the baseline is defined at 81.7\%.

\begin{table}
	\centering
	\begin{tabular}{| c | c | c | c |}

	\hline
	Classifiers & WN &  EN & SR \\
	\hline \hline
	SVM           & 81.7\% & 79.8\% & 84.4\% \\
	\hline
	Rand. Forest & 81.3\% & 80.5\% & 86.7\% \\
	\hline
	\end{tabular}
	\caption{Audio normalization of three subsets: WN - without normalization, EN - energy normalized, SR - silence removed.}
	\label{table:2}	
\end{table}

\subsection{Training results}
The first step is to choose the best normalization approach. We create three subsets of audio: without normalization (WN), energy level normalized (EN), and silence removed (SR) utterances. In order to find which technique works the best, we apply SVM and random forest to each subset.
The results are shown in table \ref{table:2}. 

From the table we can see that energy holds important information about the speaker and normalizing it worsens the predictions. In contrary, removing silences significantly improves the results in both classifiers. 
For this reason, we keep the silence removal preprocessing step in our pipeline and omit energy normalization. In addition, the random forest outperforms the SVM algorithm in the majority of cases and confirms the results of \cite{levitan2016identifying,Levitan2016Gender} on similar tasks. Random forest will be used in the rest of our experiments as the main algorithm for utterance classification.

Metadata and language features were also tested with both SVM and random forest algorithms. Each algorithm is applied to three datasets 1) time usage data, 2) show-type request distribution, and 3) language bag-of-word (BOW) features. The performance is described in table \ref{table:3}. 

We can see that time usage and show-type ratio provide very little information on who the user is. Bag-of-word model shows a prediction accuracy of 68.2\%. This result better compares to metadata, but is worse than acoustic features alone. Random forest outperforms SVM on two out of three domain of the data. For this reason, we use random forest as out main machine learning algorithm on this data. All the experiments are tested by means of cross validation on training set that we use as our development set. Due to the time complexity of deep learning algorithm, we do not use it during cross validation. Having decided on the best normalization and machine learning algorithm, we are ready to see the performance on the test data set.


\begin{table}
	\centering
	\begin{tabular}{| c | c | c | c |}

	\hline
	Classifiers & Time & Show-type & BOW \\
	\hline \hline
	SVM           & 53.4\%  &  59.9\%  &  64.7\% \\
	\hline
	Rand. Forest & 54.9\%  &  56.8\%  &  68.2\% \\
	\hline
	\end{tabular}
	\caption{Training results on meta data}
	\label{table:3}	
\end{table}

\begin{table}
	\centering
	\begin{tabular}{| c | c | c | c | c |}

	\hline
	Audio&DL&Time&Show-type&BOW\\
	\hline \hline
	86.6\% & 88.82 & 57.9\% & 55.8\% & 67.6\% \\
	\hline
	\end{tabular}
	\caption{Test results}
	\label{table:4}	
\end{table}

\subsection{Testing results}
The results on the test data are comparable to what we got during the cross validation on our train set. Table \ref{table:4} shows that the time based model provides only 57.9\% accuracy. The expectation was to get a higher score on this data set.
Our hypothesis was that TV content providers use time slots to target different age groups of their audience. Weekend mornings for animated shows and weekday nights for news are examples of such.
One reason for this might be that the commands for children shows come from parents.
We also explored show-type requests for each device to capture user interest.  
This turned out to be the least predictive data model. 
The idea was that users with children will request more child oriented content. However, insufficient data size of our show-type distribution can be attributed to such low result, and larger data set may improve the performance. This will need further investigation. Acoustic based models are the most predictive. While random forest shows improved results compared to the baseline, the deep learning method outperformed all the models and showed 88.82\% accuracy.



\begin{table}
	\centering
	\begin{tabular}{| c | c | c | c |}

	\hline
	 & Baseline & Feature & Model\\
	\hline \hline
	Accuracy & 81.7\% & 86.3\% & 90.9\% \\
	\hline
	\end{tabular}
	\caption{Feature and model level fusion results}
	\label{table:5}	
\end{table}

\begin{table}
	\centering
	\begin{tabular}{| c | c | c |}

	\hline
	 & Adult & Kid\\
	\hline \hline
	Adult & 88\% & 7\% \\
	\hline
	Kid & 12\% & 93\% \\
	\hline
	\end{tabular}
	\caption{Class confusion matrix}
	\label{table:6}	
\end{table}

\begin{table}
	\centering
	\begin{tabular}{| c | c | c | c |}

	\hline
	 & Male & Female & Kid\\
	\hline \hline
	Adult & 49\% & 39\% & 7\% \\
	\hline
	Kid & 1\% & 11\% & 93\% \\
	\hline
	\end{tabular}
	\caption{Gender confusion matrix}
	\label{table:7}	
\end{table}

\subsection{Feature and Model Fusion}
We explore feature level and model level fusion approaches to improve the results. Both techniques are known ways to combine multi-domain data. 
We concatenated features from all four data sets and trained a new random forest model. This produced somewhat of an unpredictable result. 
The accuracy of the model did not improve, and even worsened producing 86.3\% accuracy. It seems that combining all the available features into a single vector introduces noise and data sparsity problem. The acoustic model alone outperforms the feature lever fusion approach. 

For our model level fusion approach, we use ensemble algorithm AdaBoost \cite{Freund:1997:DGO:261540.261549}. AdaBoost is an adaptive model that iteratively boosts weak learner to focus on harder cases in the training dataset.  The input to this model are class probabilities from each of the five models, which are 1) random forest based acoustic, 2) time usage, 3) show-type requests, 4) bag-of-words language model, and 5) deep learning based acoustic model. The results achieved by this approach produce 90.9\% accuracy (table \ref{table:5}) manifesting in 9.2\% absolute improvement. With closer investigation of the results in table \ref{table:6}, we can see that the algorithm works better to identify children with only 7\% on false positives. However, the model produces higher error predicting adult voices as children. Table \ref{table:7} shows gender based confusion matrix. From this table we can observe that the algorithm makes the most error distinguishing female from children voices.
This comes from the fact that female voices have broader acoustic range compare to male and, as a result, they overlap with children's.


\section{Conclusion}
\label{sec:conclusion}
This work is focused on improving child and adult user classification from voice and metadata. Metadata provides additional information about user such as time, show-categories, and show-type distribution. This type of data is often ignored during research. We found that multi-domain feature level fusion did not help to improve the results. However, by combining the models using ensemble model fusion improves the performance. The system achieves 90.9\% accuracy on the task and produces state-of-the-art results. 


\section{Future work}
In our future work, we would like work to improve our model by investigating and capturing acoustic differences between female and child voices, since our current system produces the most error classifying these groups. Also, we would like to compare the performance of human engineered features and deep learning based feature representation. In addition, semi-supervised approaches have gain popularity. Data labeling is a costly and time consuming process that, for this problem, requires human annotators. Leveraging the large amount of unlabeled data can improve results even further.

\section*{Acknowledgments}
The author wishes to thank Vamsi Potluru and G. Craig Murray for their time to advice, supervise and support the project and  Comcast Research group for the opportunity to work on this project.

\newpage

\bibliography{eacl2017}
\bibliographystyle{eacl2017}

\end{document}